\documentclass[a4, 10 pt, conference]{ieeeconf} 

\IEEEoverridecommandlockouts  

\usepackage{graphics} 
\usepackage{epsfig} 
\usepackage{mathptmx} 
\usepackage{times} 
\usepackage{amsmath} 
\usepackage{amssymb}  

\usepackage{booktabs} 
\usepackage[whole]{bxcjkjatype} 
\usepackage{hyperref} 

\usepackage{multirow}

\title{\LARGE \bf
Team Flow at DRC2023: Building Common Ground and Text-based Turn-taking in a Travel Agent Spoken Dialogue System*}

\author{Ryu Hirai$^{1}$, Shinya Iizuka$^{1}$, Haruhisa Iseno$^{2}$, Ao Guo$^{1}$, Jingjing Jiang$^{1}$, \\   Atsumoto Ohashi$^{1}$ and Ryuichiro Higashinaka$^{1}$ 
\thanks{*This work was supported by a Grant-in-Aid for Scientific Research (Grant No. JP19H05692).}
\thanks{$^{1}$ Graduate School of Informatics, Nagoya University, Japan
{\tt\small
hirai.ryu.k6@s.mail.nagoya-u.ac.jp
}
} %
\thanks{$^{2}$ School of Informatics, Nagoya University, Japan
}
}

\begin{document}

\maketitle
\thispagestyle{empty}
\pagestyle{empty}

\begin{abstract}
At the Dialogue Robot Competition 2023 (DRC2023), which was held to improve the capability of dialogue robots, our team developed a system that could build common ground and take more natural turns based on user utterance texts. Our system generated queries for sightseeing spot searches using the common ground and engaged in dialogue while waiting for user comprehension.
\end{abstract}

\section{INTRODUCTION}
Multimodal dialogue systems have made improvements in providing accurate and human-like responses to users, but even so, current multimodal dialogue systems still cannot make hospitable responses. In light of this background, the Dialogue Robot Competition 2023 (DRC2023) \cite{drc2023} was organized to improve the communication capability of dialogue robots. Each team developed a dialogue system for a ``travel plan-making task'' in which the robot, acting as a travel agency clerk, helps the customer make a feasible and satisfying travel plan.

This paper reports the work of team Flow at DRC2023. To perform smooth interaction, we aimed to develop a system that could construct common ground related to sightseeing spots and could make utterances at the proper moment. Here, common ground means information shared between interlocutors, and grounding is achieved through the presentation of information by one interlocutor and its acceptance by the other \cite{clark1996using}. In most previous spoken dialogue systems, the system and the user make utterances alternately. To enable turns to be taken more naturally, we designed our system to generate responses or nod at the appropriate moment based on the textual information of utterances.

\begin{figure}[tp]
\centering
\includegraphics[clip, width=\linewidth]{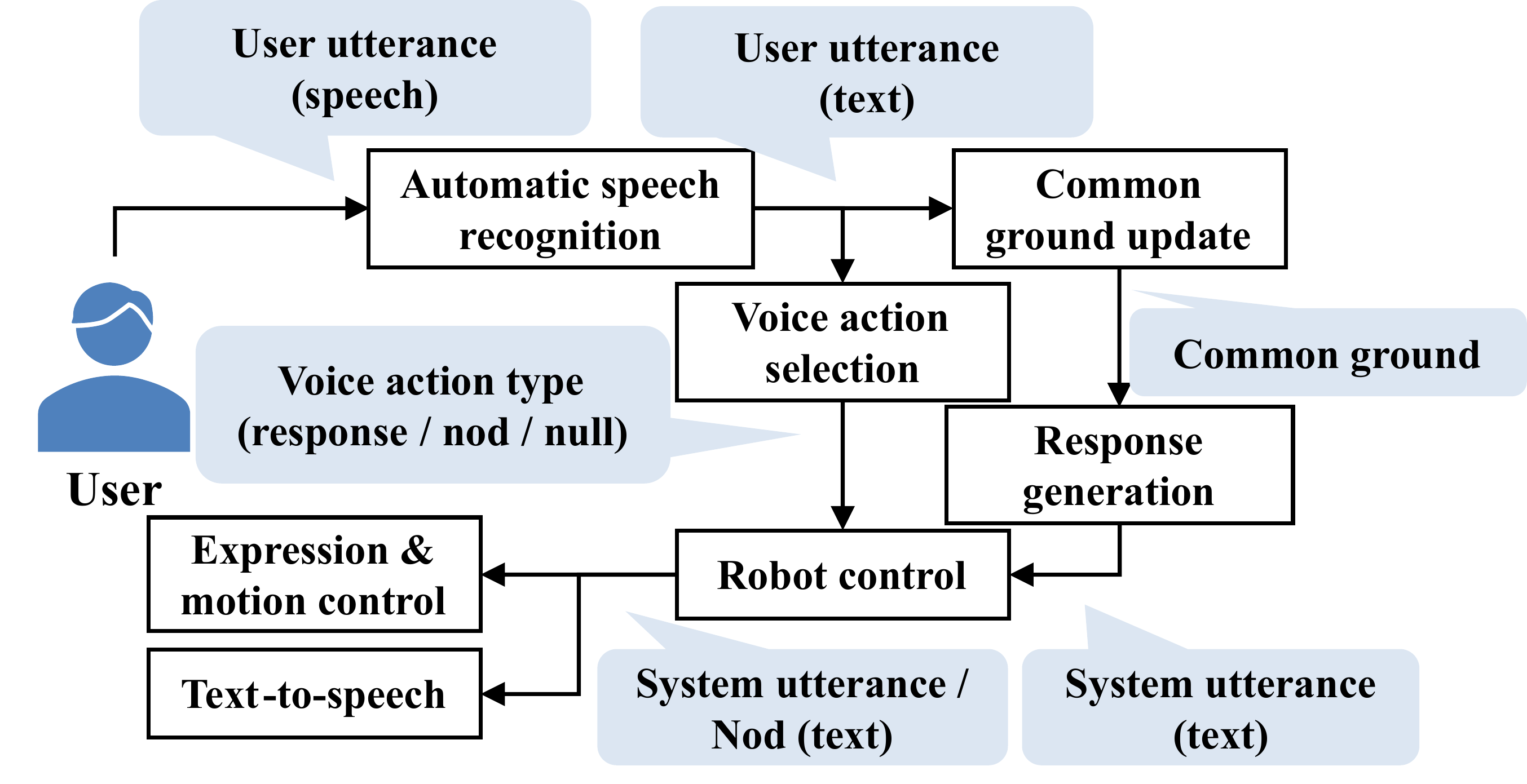}
\caption{Architecture of our system.}
\label{fig:1_overview}
\end{figure}

\section{IMPLEMENTATION}
Fig. \ref{fig:1_overview} shows the architecture of our system. The automatic speech recognition module (ASR) converts the speech into text when the system receives the user's spoken utterance. Subsequently, the user's utterance text is input to the common ground update module and the voice action selection module. The common ground update module updates the common ground using the user's utterance text, and the response generation module generates the system response based on the updated common ground. The voice action selection module outputs the voice action type the system should take by using the intermediate results of ASR. Depending on the voice action type, the generated system response is input to the text-to-speech module (TTS), or the expression and motion control module sends commands to the robot.

Regarding the implementation of the modules, we utilize the Web Speech API of Google Chrome for ASR and the Amazon Polly API for TTS. The other modules are implemented using GPT-3.5 and GPT-4 \cite{openai2023gpt}, the large language models developed by OpenAI.

The following sections describe the details of the common ground update module, response generation module, voice action selection module, and expression control and motion control module.

\begin{table*}[tp]
\centering
\caption{Dialogue example}
\label{tbl:dialogue_1515}
\begin{tabular}{lp{5cm}p{11cm}}
\toprule
Speaker & Utterance & Generated query \\
\midrule
User & Yes, I want to visit a temple where I can see beautiful autumn leaves and have a panoramic view of Kyoto. & \\
System & Alright then, & \{``Major Category'': ``Sightseeing''],``Subcategory'': ``Sightseeing --  Shrines and Temples''], ``Minor Category'': [``Buildings and Historical Sites -- Historical Buildings''], ``Other'': [``Kyoto'']\}\\
User & (Action: nod) & \\
System & With beautiful autumn leaves, & \{``Major Category'': [``Sightseeing''], ``Subcategory'': [``Sightseeing -- Shrines and Temples''], ``Minor Category'': [``Buildings and Historical Sites -- Historical Buildings''],``Other'': [``Kyoto'']\}\\
& ... & \\
User & Yes, please also tell me about amusement parks in addition to temples.& \\
System & Alright, I will try to search for amusement parks for you.& \{``Major Category'': [``Recreation''], ``Subcategory'': [``Recreation -- Theme Park'']\}\\
\bottomrule
\multicolumn{3}{p{17cm}}{The utterances and generated queries were originally in Japanese and have been translated by the authors.} \\
\end{tabular}
\end{table*}

\subsection{Common Ground Update Module}
Our system mainly records information about the user's sightseeing spot preferences, such as ``the user is interested in temples,'' as part of the common ground. As a first step of recording the common ground, we extract the user's sightseeing spot preferences from his/her utterances using GPT-4. Subsequently, we input the system response and the extracted sightseeing spot preferences into GPT-4, and GPT-4 selects the user's preferences the system has accepted. Finally, we record the selected user's preferences as common ground in a tree structure so that the topic transition of a dialogue can be reflected. 

\subsection{Response Generation Module}
Our response generation model, inspired by PaLM-SayCan \cite{brohan2023can}, first generates all possible dialogue acts (DAs) and selects the most appropriate one to respond to the user. The response generation consists of three steps. A rule-based DA generation module first generates all the DAs the system can say according to the obtained sightseeing spot information. The sightseeing spot information includes sight-seeing-spot details (e.g., descriptions, opening hours, and fees) sourced from the Rurubu Database, which the competition organizers provide. Then, a DA selection module implemented by GPT-4 selects the most appropriate DA by considering both the user's utterance and the contextual information involved in the dialogue history. Finally, a GPT-4 module is used to generate the system response according to the selected system DA.

\subsection{Voice Action Selection Module}
The voice action selection module estimates the voice action type the system should take based on the dialogue history and the ASR results of the latest user utterance. It estimates each time there is an update in the ASR results in progress and outputs one of the following types: \textbf{response}, \textbf{nod}, \textbf{nod \& backchannel}, or \textbf{none}. We implemented the voice action selection module using OpenAI's ChatGPT by few-shot learning.

\subsection{Expression Control and Motion Control Module}
The expression control and motion control module aims to select the appropriate expression and motion for each utterance generated by the response generation module. In implementing this module, we first constructed a dataset of tuples of an utterance, motion, and expression from the Travel Agency Task Dialogue Corpus \cite{Inaba2022b} using GPT-4 and then fine-tuned the Japanese BERT-v2\footnote{https://huggingface.co/cl-tohoku/bert-base-japanese-v2}, which generates motion and expression from the utterances\footnote{Our fine-tuned model could estimate expressions labeled using GPT-4 with $81\%$ accuracy and motions labeled using GPT-4 with $77\%$ accuracy.}. 


\section{RESULTS AND DISCUSSION}
In the preliminary round, the user evaluated the dialogue from two perspectives: satisfaction and travel plan. Users assessed their satisfaction on a 7-point Likert scale for nine questions. The travel plan was evaluated on the proportion of users who answered ``Yes'' to two questions: whether the user could make a travel plan and whether the plan was feasible. A total of 18 participants engaged in a dialogue, resulting in an average satisfaction rating of 2.78 and a travel plan rating of 0.28.

Table \ref{tbl:dialogue_1515} shows an example dialogue of our system. At the beginning of this dialogue, the system generated a query to search for temples based on the user's desire to visit one. Subsequently, when the user expressed a wish to go to an amusement park, the system generated a query to search for an amusement park. Our system generated queries considering the information in common ground.

As shown in Table \ref{tbl:dialogue_1515}, the system generates utterances incrementally, thus demonstrating our system's ability to engage in a dialogue while waiting for user understanding.


\section{CONCLUSION}
In this paper, we presented an overview of the system implemented by team Flow at DRC2023 and its evaluation results. We implemented a system capable of building common ground and taking more natural turns based on user utterance text. In the preliminary round, our system generated sightseeing spot search queries by managing topics using common ground and engaged in dialogue while waiting for user comprehension. In the future, we aim to explore utilizing the common ground built from system utterances and leverage multimodal information to achieve more natural turn-taking.

\addtolength{\textheight}{-12cm}


\bibliographystyle{IEEEtran}
\bibliography{sample}

\begin{thebibliography}{1}
\providecommand{\url}[1]{#1}
\csname url@rmstyle\endcsname
\providecommand{\newblock}{\relax}
\providecommand{\bibinfo}[2]{#2}
\providecommand\BIBentrySTDinterwordspacing{\spaceskip=0pt\relax}
\providecommand\BIBentryALTinterwordstretchfactor{4}
\providecommand\BIBentryALTinterwordspacing{\spaceskip=\fontdimen2\font plus
\BIBentryALTinterwordstretchfactor\fontdimen3\font minus \fontdimen4\font\relax}
\providecommand\BIBforeignlanguage[2]{{%
\expandafter\ifx\csname l@#1\endcsname\relax
\typeout{** WARNING: IEEEtran.bst: No hyphenation pattern has been}%
\typeout{** loaded for the language `#1'. Using the pattern for}%
\typeout{** the default language instead.}%
\else
\language=\csname l@#1\endcsname
\fi
#2}}

\bibitem{drc2023}
T.~Minato, R.~Higashinaka, K.~Sakai, T.~Funayama, H.~Nishizaki, and T.~Nagai, ``{Overview of Dialogue Robot Competition 2023},'' in \emph{Proceedings of the Dialogue Robot Competition 2023}, 2023.

\bibitem{clark1996using}
H.~H. Clark, \emph{{Using Language}}.\hskip 1em plus 0.5em minus 0.4em\relax Cambridge university press, 1996.

\bibitem{openai2023gpt}
OpenAI, ``{GPT-4 Technical Report},'' \emph{arXiv preprint arXiv:2303.08774}, 2023.

\bibitem{brohan2023can}
A.~Brohan, Y.~Chebotar, C.~Finn, K.~Hausman, A.~Herzog, D.~Ho, J.~Ibarz, A.~Irpan, E.~Jang, R.~Julian, \emph{et~al.}, ``Do as {I} can, not as {I} say: {G}rounding language in robotic affordances,'' in \emph{Conference on Robot Learning}.\hskip 1em plus 0.5em minus 0.4em\relax PMLR, 2023, pp. 287--318.

\bibitem{Inaba2022b}
M.~Inaba, Y.~Chiba, R.~Higashinaka, K.~Komatani, Y.~Miyao, and T.~Nagai, ``{Collection and Analysis of Travel Agency Task Dialogues with Age-Diverse Speakers},'' in \emph{Proceedings of the Thirteenth Language Resources and Evaluation Conference}, 2022, pp. 5759--5767.

\end{thebibliography}

\end{document}